\documentclass[runningheads]{llncs}
\usepackage{eccv}
\usepackage{eccvabbrv}
\usepackage{graphicx}
\usepackage{booktabs}
\usepackage{multirow}
\usepackage{amsmath}
\usepackage{amssymb}
\usepackage{enumitem}
\usepackage{subcaption}
\usepackage{xcolor}
\usepackage{microtype}
\usepackage{placeins}

\usepackage[pagebackref,breaklinks,colorlinks,citecolor=eccvblue]{hyperref}

\begin{document}

\title{The Plan, Not the Decoder: Diagnosing and Repairing Compositional Failure in Reasoning-Augmented Text-to-Image Generation}
\titlerunning{The Plan, Not the Decoder}
\author{Ashritha Gonuguntla}
\authorrunning{A. Gonuguntla}
\institute{Carnegie Mellon University \\ \email{agonugun@andrew.cmu.edu}}

\maketitle

\begin{abstract}
Reasoning-augmented text-to-image (T2I) models such as GoT-R1 emit an explicit textual plan---objects, attributes, and bounding boxes---before generating image tokens. When such models fail compositionally, is the plan wrong, or is the plan right and the decoder unfaithful? We answer this with a controlled intervention study on GoT-R1-1B over T2I-CompBench++, scored geometrically with an open-vocabulary detector after showing that a widely used VQA-based spatial metric is blind to layout: swapping the two bounding boxes inside the chain flips the generated layout (detector accuracy $0.75\!\to\!0.48$, $p{<}10^{-3}$) yet \emph{raises} the VQA score. Under sound measurement, the decoder is a strikingly faithful plan executor: 94\% of generated layouts realize the planned relation, object--box binding survives reordering of the plan's object segments, and---contrary to a co-adaptation hypothesis our own earlier experiments appeared to support---injected external plans are executed \emph{better} than the planner's own: correct plans with clean, well-separated boxes improve spatial accuracy by +13.3 points (paired $p{=}10^{-4}$), regardless of whether their text is terse or verbose, and regardless of their likelihood under the planner (a 5$\times$ NLL gap with no execution cost). Planner-mimicking plans that reuse the planner's own overlapping box statistics gain nothing, and a causal intervention---\emph{geometric plan repair}, which keeps the planner's objects and prose and rewrites only its box geometry---recovers most of the replacement gain ($+10.7$, $p{<}10^{-4}$) at zero content cost, decomposing into sign correction (dominant) plus a modest separation threshold ($\approx$0.2 normalized margin, beyond which more buys nothing). The bottleneck is therefore the \emph{plan}: the planner writes wrong relations for phrasing-dependent reasons (98\% vs.\ 54\% planning accuracy on semantically identical layouts, a raster-order bias we isolate with a mention-order control) and cluttered geometry the decoder faithfully reproduces. Practical consequences follow immediately: symbolic plan verification with resampling (+5.0, $p{<}10^{-3}$) and minimal in-place plan repair (+6.0, $p{=}.02$) are training-free wins, and modular planner--decoder designs are viable---provided the plan is internally consistent, since box--text contradictions induce object duplication and identity fusion. We release our plan-fidelity evaluation protocol, all plans, and 12k generated images.
\end{abstract}

\section{Introduction}
\label{sec:intro}

Compositional prompts remain a reliable way to break modern text-to-image models: asked for ``a cat on the left of a dog,'' a model may draw both animals and reverse the relation. A prominent recent response is to make the model \emph{reason before it draws}: generation chain-of-thought (GoT) models~\cite{got, gotr1} emit a textual plan---object identities, attributes, bounding-box coordinates---between the prompt and the image tokens, and train the planner with reinforcement learning against rewards that include plan-to-image alignment.

This design creates a two-stage failure surface: the plan can be wrong, or the plan can be right and poorly executed. Benchmark scores conflate the two, and recent benchmark-level studies report that unified models' reasoning is not reliably reflected in their pixels~\cite{ureason, realunify}. GoT-style plans make the question directly decidable, per prompt: the plan states machine-checkable constraints that can be compared against both the prompt and the rendered image. We decide it for GoT-R1-1B~\cite{gotr1} with a controlled intervention program---verify, repair, replace, and corrupt the plan---in which every condition shares hardware, seeds, decoding settings, and conditioning path with an unperturbed control, and every headline delta carries a paired permutation test. Our contributions:

\begin{enumerate}[itemsep=2pt,topsep=2pt]
    \item \textbf{Metric validation, with a built-in ground truth.} Swapping the two boxes inside the model's own chain is a manipulation whose effect on layout we can verify visually; the standard BLIP-VQA spatial question scores this corruption as an \emph{improvement}, while detector-based geometric scoring correctly reports a 27-point collapse. VQA-style relation questions should not be used to evaluate spatial execution (\S\ref{sec:metrics-validation}); all our spatial conclusions use the detector.
    \item \textbf{The decoder is a faithful executor.} On unperturbed chains, 94\% of generated layouts realize the planned relation (planned-vs-detected box IoU 0.75); reversing the order of the plan's object segments leaves prompt accuracy unchanged while ``followed the literal plan'' collapses---the decoder binds objects to boxes semantically, not positionally; and ablating the chain collapses object presence, confirming the plan is the primary conditioning channel (\S\ref{sec:perturb}).
    \item \textbf{External plans transfer---geometry, not style or familiarity, decides.} A $3\times2$ ``oracle dial''---crossing three text styles (terse, verbose, or planner-mimicking) with two box distributions (clean vs.\ planner-statistics)---shows correct injected plans with clean, well-separated boxes beat the model's own plans by +13.3 points ($p{=}10^{-4}$), verbose or terse alike; plans reusing the planner's overlapping box statistics gain nothing; and planner likelihood is irrelevant (5$\times$ NLL gap, no execution cost). This contradicts a planner--decoder co-adaptation account---including our own earlier finding, whose reversal we document and explain (\S\ref{sec:dial}, \S\ref{sec:revisiting}).
    \item \textbf{The planner is the bottleneck, and it can be patched at inference time.} The planner writes invalid relations for roughly one prompt in five, with a striking phrasing dependence---98\% vs.\ 54\% planning accuracy on semantically identical layouts, isolated by a mention-order control as a raster-order bias (\S\ref{sec:gap})---and emits cluttered, overlapping boxes that cap execution quality. Training-free fixes follow the diagnosis: symbolic verification with resampling (+5.0, $p{<}10^{-3}$) and minimal in-place repair (+6.0, $p{=}.02$) (\S\ref{sec:interventions}).
\end{enumerate}

The practical moral: reasoning-augmented decoders are better plan executors than their planners are plan writers. Fixing, or outright replacing, the plan at inference time is cheap and effective---the one hazard is internal inconsistency, since plans whose boxes contradict their text induce duplication and identity fusion.

\section{Related Work}
\label{sec:related}

\paragraph{Reasoning-augmented and unified image generation.}
GoT~\cite{got} and GoT-R1~\cite{gotr1} generate a semantic-spatial plan before image tokens, with GoT-R1 adding RL over MLLM-based rewards; T2I-R1~\cite{t2ir1}, ReasonGen-R1, and BAGEL~\cite{bagel} pursue related recipes. At benchmark scale, UReason~\cite{ureason} and RealUnify~\cite{realunify} report that unified models' reasoning is unreliably transferred to pixels, and Dong~\etal~\cite{gta} name the reasoning-right/action-wrong pattern an execution gap in GUI agents. Our per-prompt, interventional accounting reaches a different conclusion for GoT-R1: under geometric scoring the decoder executes plans faithfully, and apparent reasoning--pixel misalignment largely reflects planner errors and metric artifacts.

\paragraph{Perturbation-based faithfulness analysis.}
Our methodology ports the perturbation paradigm developed for LLM chain-of-thought faithfulness~\cite{lanham2023faithfulness, turpin2023unfaithful} to visual generation; related interventions exist for multimodal \emph{understanding}~\cite{visualthinking2025, medvlmperturb2025} but, to our knowledge, not for the plans of reasoning-augmented image generators.

\paragraph{Inference-time verification and self-correction.}
A dense line of work verifies or corrects the \emph{image}: SLD detects and edits objects post hoc~\cite{sld}; PARM scores intermediate generations~\cite{parm}; Reflect-DiT~\cite{reflectdit} and iterative-refinement pipelines~\cite{iterrefine} critique and regenerate; EPIC compiles prompts into symbolic predicates and routes failed images to editing or resampling~\cite{epic}. Our verifier applies the symbolic-constraint idea one stage earlier, to the textual plan, where a failed check costs a $\sim$5\,s chain resample instead of an $\sim$18\,s decode; our results say the two loci catch different errors (plan content vs.\ residual execution noise) and compose naturally. Concurrent work verifies reasoning trajectories during \emph{training}~\cite{clvr} or plugs decoupled planners into generators with reinforcement learning~\cite{thinkstrokes}; notably, the authors of~\cite{thinkstrokes} report that an untuned external planner \emph{degrades} their base model, whereas we find external plans excel for a plan-trained decoder---evidence the transferability of plans is a property of the decoder's training regime, not of planning per se.

\paragraph{Layout conditioning and conflicting conditions.}
LayoutBench reports that layout-conditioned diffusion models collapse on out-of-distribution ground-truth layouts~\cite{layoutbench}, and follow-ups attribute layout-to-image failures to conditioning distribution shift~\cite{laysyn}; misaligned condition pairs are known to destabilize diffusion models~\cite{decomposerealign}. For an RL-trained autoregressive plan-follower we find the opposite of the distribution-shift account: alien layouts execute \emph{better} than familiar ones, and the destabilizing factor is internal box--text contradiction, whose failure signature (duplication, identity fusion) we document. Compositionality bottlenecks in T2I text encoders~\cite{kamath2023text, doesclipbind, whatsup, spright, lessfromtext} motivate explicit plans as the carrier of binding structure; mention-order biases in generation~\cite{visor, otsbench} anticipate the planner-stage raster bias we isolate.

\section{Experimental Setup}
\label{sec:setup}

\paragraph{Model.} GoT-R1-1B~\cite{gotr1} augments Janus-Pro-1B~\cite{januspro} with a generation chain-of-thought stage: given prompt $p$, it emits a plan containing delimited object spans (e.g., ``red car'') each followed by a delimited bounding box $(x_1,y_1),(x_2,y_2)$ with coordinates in $[0,1000)$, then generates 576 image tokens conditioned on prompt and plan (384px output). Unless stated otherwise: CFG weight 5.0, temperature 1.0, seed 1000, single seed per prompt.

\paragraph{Data.} T2I-CompBench++~\cite{t2icompbench} provides the two compositional validation subsets we evaluate: color (attribute binding) and spatial (2D relations), 300 prompts each. We use the validation subsets for compute reasons and emphasize paired, same-rig comparisons over absolute rankings. We note for precision that T2I-CompBench++'s \emph{official} spatial metric is detector-based (UniDet~\cite{unidet}), not VQA-based; the BLIP-VQA judge is the benchmark's official metric for attribute binding, and VQA-style relation questions are a common practice in the surrounding literature rather than this benchmark's prescribed spatial protocol. Our detector-based scoring is therefore methodologically aligned with the benchmark's own spatial evaluation, and the artifact we document in \S\ref{sec:metrics-validation} concerns VQA-style spatial scoring as practiced, not T2I-CompBench++'s prescription.

\paragraph{Experimental controls.} All conditions in \S\ref{sec:interventions}--\ref{sec:perturb} were generated in a single hardware/software environment with identical seeds and decoding settings, and every injected-chain condition shares one conditioning path (chain text re-tokenized before decoding) with an unperturbed control generated through that same path. Headline deltas are tested with paired sign-flip permutation tests over prompts ($10^4$ resamples).

\paragraph{Metrics.} \emph{Spatial:} we detect both objects with OWLv2~\cite{owlv2} and score (i) \textbf{relation correctness} (detected centers satisfy the prompt's relation), (ii) \textbf{plan-execution fidelity} (planned-vs-detected box IoU per object, and whether the detected layout realizes the \emph{planned} relation). \S\ref{sec:metrics-validation} validates this choice against VQA-based scoring. \emph{Color:} the diagnostic BLIP-VQA suite~\cite{blip} (object presence; binary and open-ended color questions per object), where VQA is reliable. \emph{Plan level:} we parse boxes from the chain and check the prompt's relation against box centers. CLIPScore~\cite{clipscore} carries no usable signal about layout once object presence is controlled: among images containing both named objects, those realizing the prompt's relation score 0.3287 against 0.3239 for those that do not ($r{=}0.066$, $n{=}272$, n.s.). A weak positive association over the full split ($r{=}0.131$) reflects whether the objects are present at all rather than how they are arranged, and disappears under that control. CLIPScore is therefore not used.

\section{The Planner Errs; Sound Metrics See It}
\label{sec:gap}

\paragraph{Plan validity.} Parsing GoT-R1's own chains, 79.7\% of the 192 asymmetric-relation spatial prompts with two checkable boxes are planned consistently with the prompt's relation; under the stricter verifier of \S\ref{sec:verify} (which also rejects chains with fewer than two parsed objects), 76.3\% of spatial and 77.7\% of color chains pass on the first sample. The planner errs on roughly one prompt in five.

\paragraph{Planner asymmetry: a raster-order bias, isolated by a mention-order control.} Planning accuracy is far from uniform across relations. To separate a directional prior from mention-order bias~\cite{otsbench, visor}, we re-plan every asymmetric prompt in its semantically equivalent inverted phrasing (``X on the left of Y'' $\leftrightarrow$ ``Y on the right of X''; 3 seeds, chains only). Accuracy tracks the \emph{relation word}, not the objects or their order: left 97.8\%/98.0\% (original/inverted), right 55.1\%/53.9\%, top 89.5\%/87.5\%, bottom 73.7\%/77.9\%. Two phrasings of the identical layout differ by 44 points of planning accuracy, and the deficit pattern---right and bottom hard, left and top easy---is what a raster-order prior predicts: the planner prefers to place the first-mentioned entity at the top-left origin. This extends order-to-space bias~\cite{otsbench} to the explicit planning stage, where it is directly observable in box coordinates.

\subsection{Metric validation: VQA cannot see layout; a detector can}
\label{sec:metrics-validation}

Before scoring execution we validate the ruler, using a manipulation with known ground truth: \textbf{box swap}. Exchanging the two bounding boxes inside the model's own chain (text untouched) demonstrably flips generated layouts (Figure~\ref{fig:qualitative}, left/middle). Under OWLv2 geometric scoring, prompt-relation accuracy collapses from 0.750 to 0.477 ($-27.3$ points, 95\% CI $[-34.3, -20.0]$, $p{=}10^{-4}$)---as it must. Under the BLIP-VQA spatial question (``Is the X on the left of the Y?''), the same corruption \emph{raises} the measured score (0.618 $\to$ 0.681): the VQA judge is yes-biased and largely layout-insensitive, echoing known VLM spatial weaknesses~\cite{whatsup}. A five-rater human study over 240 stratified images confirms the choice of ruler: detector verdicts agree with the human majority on 81\% of items and reproduce the human ordering over all conditions, while BLIP-VQA sits at 57\% (near chance); a modern instruct VLM judge (Qwen2.5-VL~\cite{qwen25vl}) also passes the corruption test and reaches 84\% item agreement, but runs systematically conservative in aggregate and inherits the same brittle prompt parser as BLIP (covering 207/300 prompts vs.\ the detector's 300). Detector-based scoring is not assumption-free either---open-vocabulary detection quality bounds it~\cite{saneval}---but among the three judges it is the one that passes the corruption test, best matches humans, and covers all prompts. All spatial results below therefore use detector-based scoring; we flag the BLIP artifact because VQA-style spatial metrics remain in wide use.

\subsection{The decoder executes faithfully}
On unperturbed chains (the control condition), 95.7\%/95.0\% of planned objects are detected, planned-vs-detected box IoU is 0.747, and \textbf{93.8\% of generated layouts realize the planned relation}. Given plans, this decoder draws them. Prompt-level spatial accuracy on the same images is 0.750---close to the 79.7\% plan-validity rate once detection noise is accounted for. The residual compositional failures are dominated by planner errors, not execution errors.

\section{Intervening on the Plan}
\label{sec:interventions}

If the plan is the bottleneck, improving it at inference time should transfer directly to images. We test three escalating interventions---verify, repair, replace---against matched controls (Table~\ref{tab:interventions}).

\begin{table}[t]
\centering
\small
\resizebox{\textwidth}{!}{%
\begin{tabular}{@{}llccc@{}}
\toprule
Condition & vs.\ control & Relation$\checkmark$ $\uparrow$ & $\Delta$ [95\% CI] & $p$ \\
\midrule
Unperturbed control (own chains) & --- & 0.750 & --- & --- \\
Native-path baseline (fresh chains) & --- & 0.773 & --- & --- \\
\midrule
Verify-then-generate ($K{=}5$) & native baseline & \textbf{0.823} & $+5.0$ $[+2.3, +8.0]$ & $.0007$ \\
Minimal in-place repair & control & \textbf{0.810} & $+6.0$ $[+1.0, +11.0]$ & $.021$ \\
Geometric plan repair ($m{=}0.2$) & control & \textbf{0.857} & $+10.7$ $[+6.7, +14.7]$ & $<.0001$ \\
\midrule
Oracle: minimal text, clean boxes & control & \textbf{0.883} & $+13.3$ $[+8.3, +18.3]$ & $.0001$ \\
Oracle: verbose text, clean boxes & control & \textbf{0.870} & $+12.0$ $[+6.7, +17.3]$ & $.0001$ \\
Oracle: minimal text, planner-stat boxes & control & 0.780 & $+3.0$ $[-3.0, +8.7]$ & $.37$ \\
Oracle: verbose text, planner-stat boxes & control & 0.803 & $+5.3$ $[0.0, +10.7]$ & $.062$ \\
Oracle: planner-mimic (donor chains) & control & 0.740 & $-1.0$ $[-7.0, +5.0]$ & $.83$ \\
\bottomrule
\end{tabular}}%
\caption{\textbf{The intervention ladder} (OWLv2 relation correctness, \texttt{spatial\_val}, $n{=}300$ per condition, paired sign-flip permutation tests). Verification and repair give significant training-free gains; \emph{replacing} the plan outright helps most---provided the boxes are clean and well-separated. Text style (minimal vs.\ verbose) is irrelevant within each box type; plans mimicking the planner's own style and box statistics gain nothing.}
\label{tab:interventions}
\end{table}

\subsection{Verify-then-generate: symbolic plan verification with rejection sampling}
\label{sec:verify}

Before any image decoding, we parse the chain and check it against constraints parsed from the prompt (box-center inequality for the stated relation; presence of each prompt color; $\geq$2 parsed objects). On failure, the chain---never the image---is resampled at temperature $1.0 + 0.1k$ on retry $k$, up to $K{=}5$; if all fail, the base chain is used. Verification is string parsing and geometry (microseconds); a chain resample costs $\sim$5\,s vs.\ $\sim$18\,s per decode.

\paragraph{Retry statistics.} Spatial: 229/300 chains pass first-try; resampling recovers 48 of the remaining 71 (68\%), leaving 23 fall-backs (7.7\%). Color: 233/300 first-try, 56/67 recovered (84\%), 11 fall-backs. Verified prompts require 1.26/1.32 chain samples on average---overhead is a fraction of one chain generation per prompt.

\paragraph{Results.} Against its path-matched fresh-chain baseline, verification improves detector relation accuracy 0.773 $\to$ 0.823 ($+5.0$, $p{=}.0007$); on color prompts (BLIP diagnostics), combined accuracy improves 0.864 $\to$ 0.897. The gain is bounded by what verification can do: it converts invalid plans into valid ones and leaves execution untouched.

\subsection{Minimal in-place repair}
\label{sec:repair}

Where verification resamples, repair \emph{edits}: when a spatial plan fails the check, we swap the two box coordinate strings (the smallest content fix; prose untouched); when a color plan omits a required color, we insert it into the object's span. Repaired chains are nearly indistinguishable from the planner's own by likelihood (mean NLL 0.553 vs.\ 0.519 nats/token). Across 300 spatial prompts, 71 plans were invalid and 40 were repairable; the condition improves relation accuracy 0.750 $\to$ 0.810 ($+6.0$, $p{=}.021$). Repair slightly outperforms verification per unit compute (no resampling loop) and, unlike resampling, is deterministic.

\subsection{Replacing the plan: the oracle dial}
\label{sec:dial}

Finally we discard the planner's content entirely and inject constructed, correct plans, along a dial that unconfounds three properties: \emph{text style} (terse one-sentence; verbose multi-sentence with direction-correct descriptions; or \emph{planner-mimic}---a real donor chain from the planner's own outputs with object names substituted, preserving register, length, and box statistics exactly), and \emph{box statistics} (clean: large, well-separated, fixed margins; planner-stat: sampled from the planner's own box distribution, overlap included).

Table~\ref{tab:interventions} (bottom) shows a sharp dissociation. \textbf{Box geometry decides; style does not.} With clean boxes, injected plans beat the planner's own by $+13.3$ (terse) and $+12.0$ (verbose) points; with planner-statistics boxes the same texts gain nothing significant; and planner-mimic plans, maximally familiar in every respect, sit exactly at the control. \textbf{Familiarity is irrelevant:} planner NLL rises steeply for every injected plan---own chains (0.52) $<$ planner-mimic (1.53) $<$ terse oracle (2.61) $<$ verbose oracle (3.16)---yet execution quality \emph{anti-correlates} with likelihood across this range. The decoder is not co-adapted to its planner's distribution; it is a general plan executor whose output quality tracks the geometric quality of the plan it is handed.

\paragraph{The oracle gain is mediated by geometry---and a causal test decomposes it.} A skeptic may object that clean, well-separated boxes define an easier target for generation and detection alike. Observationally this is right: binning the \emph{control} condition's own plans by normalized center margin shows accuracy rising from 0.49 (margin $<0.1$) to 1.00 ($\geq 0.45$), the oracle's boxes (mean margin 0.54 vs.\ the planner's 0.26) land in the top band, and at matched margins injected plans hold no advantage over the model's own. To make the geometry claim causal we intervene directly: \textbf{geometric plan repair} keeps the planner's own objects and prose and rewrites only the two boxes, preserving their sizes, so that centers satisfy the prompt relation at a target margin. Repairing to margin 0.2 improves overall relation accuracy by $+10.7$ points ($p{<}10^{-4}$; Table~\ref{tab:interventions})---the strongest training-free intervention we test, at zero content cost. Decomposing on the 192 modified (asymmetric) prompts: the gain is dominated by \emph{sign correction} on the 39 originally-invalid plans ($+64.1$ points, $p{<}10^{-4}$), while pure margin widening on already-valid plans contributes a small positive effect ($+3$--$5$ points, individually below significance), and pushing the target margin beyond 0.2 buys nothing (0.857/0.853/0.850 at targets 0.2/0.4/0.6). The causal picture is therefore a \textbf{threshold, not a gradient}: what matters is that the plan's relation is \emph{right} and its boxes are \emph{separated enough} ($\approx$0.2 normalized margin); the observational margin curve above partly reflects prompt-difficulty confounds beyond the causal margin effect. The planner's cluttered, low-margin, occasionally sign-flipped boxes are the system's ceiling---and a three-line geometric edit at inference time recovers most of what full plan replacement achieves.

\subsection{What breaks execution: internal contradiction}
\label{sec:contradiction}

\begin{figure}[t]
\centering
\begin{subfigure}{0.32\textwidth}
  \includegraphics[width=\linewidth]{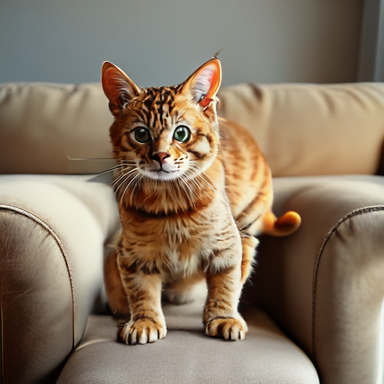}
  \caption{Own chain}
\end{subfigure}\hfill
\begin{subfigure}{0.32\textwidth}
  \includegraphics[width=\linewidth]{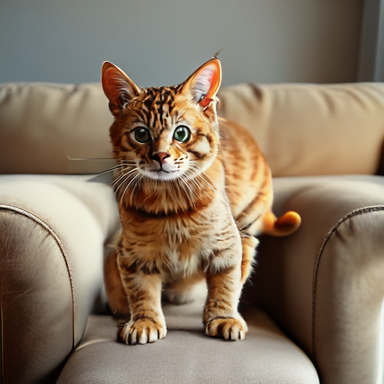}
  \caption{Verified chain}
\end{subfigure}\hfill
\begin{subfigure}{0.32\textwidth}
  \includegraphics[width=\linewidth]{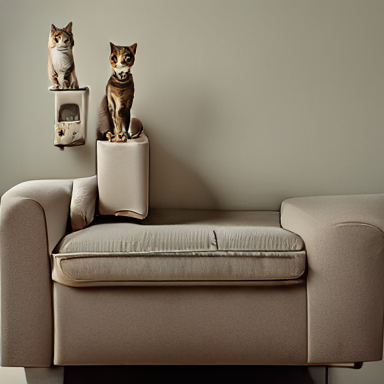}
  \caption{Contradictory plan}
\end{subfigure}
\caption{\textbf{``A cat on the top of a sofa.''} Verified chains preserve the baseline's coherence. Internally \emph{contradictory} plans (boxes disagreeing with prose, as in box-swap and our legacy oracle construction) produce stacked, duplicated objects---the failure signature of conflicting conditioning, not of plan novelty.}
\label{fig:qualitative}
\end{figure}

One manipulation reliably damages generation beyond its geometric content: plans whose components \emph{disagree}. Box-swapped chains (boxes flipped, prose still describing the original layout) not only flip layouts but produce characteristic object duplication and identity fusion (Figure~\ref{fig:qualitative}, right: a man--giraffe hybrid); an earlier, buggy version of our verbose template whose description clause contradicted its boxes reproduced the same degradation signature. Attribute fusion under conflicting conditioning has been observed for external condition pairs in diffusion models~\cite{decomposerealign}; we find it arises \emph{within} a single self-generated plan whose channels conflict. Practical rule: plan editors must keep boxes and prose consistent---or strip the prose.

\section{What Does the Decoder Read? Perturbing the Plan}
\label{sec:perturb}

To identify which chain components the decoder consumes, we apply controlled perturbations to pre-generated chains and regenerate through an identical pipeline against an unperturbed control, porting the perturbation-faithfulness methodology of LLM chain-of-thought analysis~\cite{lanham2023faithfulness, turpin2023unfaithful} to visual generation.

\begin{table}[t]
\centering
\small
\resizebox{\textwidth}{!}{%
\begin{tabular}{@{}lcccc@{}}
\toprule
Condition (spatial) & Relation$\checkmark$ $\uparrow$ & Plan IoU & Followed plan & Obj.\ detected \\
\midrule
Unperturbed control & 0.750 & 0.747 & 0.938 & .957/.950 \\
Box swap & 0.477 & 0.629$^{\dagger}$ & 0.755$^{\dagger}$ & .947/.917 \\
Boxes removed & 0.600 & --- & --- & .903/.867 \\
Object segments reversed & 0.747 & 0.282 & 0.096 & .957/.947 \\
Chain ablated & 0.453 & --- & --- & .800/.723 \\
\bottomrule
\end{tabular}}%
\caption{\textbf{Spatial chain perturbations} (OWLv2; $n{=}300$; $^{\dagger}$fidelity scored against the \emph{swapped} plan). The decoder reads boxes (swap flips layouts, $-27.3$, $p{=}10^{-4}$; removal costs 15 points), reads them \emph{semantically} (reversing segment order changes nothing about the image---``followed the literal reordered plan'' collapses to 0.096 while prompt accuracy is untouched), and depends on the chain for object identity itself (ablation: $-29.7$ relation, $-16$ points object detection).}
\label{tab:perturb-spatial}
\end{table}

\begin{table}[t]
\centering
\small
\begin{tabular}{@{}lcccc@{}}
\toprule
Condition (color subset) & Presence $\uparrow$ & Color Y/N $\uparrow$ & Color match $\uparrow$ & Combined $\uparrow$ \\
\midrule
Unperturbed control & 0.9219 & 0.9160 & 0.7884 & 0.8564 \\
Attribute swap in chain & 0.9238 & 0.7751 & \textbf{0.5605} & 0.7233 \\
Boxes removed from chain & 0.8320 & 0.8431 & 0.7370 & 0.7386 \\
Chain ablated entirely & \textbf{0.6934} & 0.6816 & 0.6146 & 0.5234 \\
\bottomrule
\end{tabular}
\caption{\textbf{Color chain perturbations} (diagnostic BLIP-VQA, reliable for color; $n{=}256$ of 300 prompts, the remainder not parsed by the benchmark's question generator). Swapping the two colors inside the chain---prompt untouched---transfers the swap to the image: color match falls 23 points with object presence unchanged. The chain dominates the prompt for attribute binding, though it does not fully override it.}
\label{tab:perturb-color}
\end{table}

Three findings. \textbf{(i) The decoder reads geometry and attributes from the chain, not the prompt.} Swapped boxes flip layouts (Table~\ref{tab:perturb-spatial}); swapped colors recolor objects (color match $0.788\!\to\!0.561$, presence unchanged; Table~\ref{tab:perturb-color}). When chain and prompt disagree, the chain dominates. \textbf{(ii) Binding is semantic, not positional.} Reversing the order of the plan's object segments leaves every image-level metric at control levels while agreement with the \emph{literal} reordered plan collapses to 0.096: the decoder re-associates each object with its box by meaning and renders the same scene. \textbf{(iii) The chain is the primary conditioning channel.} Ablating it collapses object detection by 16--23 points on both subsets---after plan-conditioned RL training, the prompt alone no longer reliably specifies even which objects to draw.

\paragraph{Mechanism probes (exploratory).} Aggregate cross-attention from image tokens to the plan's box tokens is small ($\approx$0.7\% of conditioning attention) and statistically flat across own-chain, mimic, and verbose-oracle conditions, with at most weak correlation to per-prompt success ($r{=}0.17$ on the control, $\approx$0 elsewhere). Whatever computation reads the boxes, it is not visible as bulk attention reallocation; identifying it (specific heads, coordinate-token representations) is future work.

\section{Robustness: Seeds and Scale}
\label{sec:robustness}

\paragraph{Seeds.} Regenerating the headline conditions with two additional generation seeds leaves every conclusion intact: control $0.751 \pm 0.005$, minimal repair $0.834 \pm 0.024$, geometric plan repair $0.849 \pm 0.007$, clean-box oracle $0.890 \pm 0.009$ (mean $\pm$ std over three seeds). The ladder's ordering holds in every individual seed.

\paragraph{Scale: GoT-R1-7B.} Replicating the core protocol on GoT-R1-7B confirms every qualitative finding and adds a scaling observation. The 7B decoder is equally faithful (followed-plan 0.92, IoU 0.76); box swap collapses detector accuracy $0.863 \to 0.433$ ($-43$ points, $p{<}10^{-4}$), re-validating the metric at scale; and every intervention still improves over the control (verify 0.887, minimal repair 0.880, margin repair 0.883, clean-box oracle 0.927, $p{=}.001$; the fine ordering among the near-tied repair variants differs from 1B, as expected close to ceiling). The interesting difference is compression: the 7B \emph{planner} is much better---89.3\% of chains pass verification first-try vs.\ 76.3\% at 1B---so the baseline rises to 0.863 and the plan-repair gains shrink toward noise ($+1.7$ to $+2.3$, individually n.s.; only full replacement stays significant at $+6.3$). This is precisely what the thesis predicts: \textbf{scale buys a better planner, not a different executor}, and inference-time plan repair matters most where planners are weakest. (The planner-mimic condition degenerates at 7B for a stimulus reason---name substitution leaves donor-specific prose, creating exactly the internal contradictions of \S\ref{sec:contradiction}, which the more prose-faithful 7B decoder renders; we therefore do not interpret that cell.)

\section{Revisiting Our Earlier Finding, and Why It Reversed}
\label{sec:revisiting}

An earlier version of this project (course-project experiments, different hardware and library versions) found the opposite headline: injected oracle plans \emph{degraded} every metric, which we interpreted as planner--decoder co-adaptation. Three defects, uncovered by an adversarial audit of our own pipeline, explain the reversal. (1)~\emph{Corrupted stimuli:} our original oracle templates contained ungrammatical and direction-contradicting clauses---internally inconsistent plans of exactly the kind \S\ref{sec:contradiction} shows are genuinely harmful. (2)~\emph{Metric artifact:} spatial claims rested partly on VQA-based scoring that \S\ref{sec:metrics-validation} shows is layout-insensitive. (3)~\emph{Underpowered deltas:} the legacy degradation ($-4.9$ points, $n{=}300$, unpaired) sits below the minimal detectable difference of that design. With corrected templates, geometric scoring, matched controls, and paired tests, the effect reverses sign and grows in magnitude ($-4.9 \to +13.3$, $2.7\times$ larger). We report this trajectory because the interim conclusion is one the literature could plausibly adopt elsewhere: apparent ``plans don't transfer'' results deserve scrutiny of stimulus integrity and metric validity before a co-adaptation interpretation.

\section{Why Explicit Plans in the First Place}
\label{sec:support}

Two context points frame why a plan interface is the right object of study. First, no model family solves compositional generation outright: in our companion cross-family experiments, strong diffusion models lead attribute binding while GoT-R1 leads spatial grounding, and prompt-level rewrites are fragile---explicit structure and iterative decoding address different failure modes. Second, the structure has to be carried \emph{somewhere}: probing the text encoders of CLIP, SDXL, FLUX, and SD3 with swap-margin diagnostics yields near-zero binding margins and 0\% linear recovery of the ordered spatial triple, echoing known compositionality bottlenecks in contrastive encoders~\cite{kamath2023text, doesclipbind, lessfromtext}. If a global embedding barely encodes bindings, they must be carried explicitly---by a plan---which is exactly what makes the plan-execution interface the decisive locus we study here.

\section{Discussion and Design Implications}
\label{sec:discussion}

\textbf{(1) Patch the plan, not the decoder.} The intervention ladder is monotone in how much of the plan is fixed: verify $+5$, repair $+6$, replace-with-clean-geometry $+13$. All are training-free, and plan-level checks compose with image-level verification~\cite{epic, sld, parm}: plan checks catch content errors before the expensive decode; image checks catch residual execution noise after it. \textbf{(2) Modularity is viable.} Contrary to co-adaptation concerns, external planners---LLM layout generators~\cite{layoutgpt, rpg}, human edits, symbolic planners---can drive this decoder \emph{better than its own planner}, with two rules: keep boxes clean and well-separated, and keep prose consistent with boxes (or omit it). \textbf{(3) Train the planner's geometry.} The decoder faithfully reproduces the planner's cluttered boxes; the highest-leverage training target is therefore the planner's box statistics (separation, overlap), not decoder capacity. The raster-order bias suggests simple data-side fixes (order-balanced planning data). \textbf{(4) Measure geometrically.} VQA-based spatial metrics can invert the sign of a 27-point effect; detector-based plan-fidelity scoring---which we release---is cheap and validated by construction.

\section{Limitations}
\label{sec:limitations}

One model family (GoT-R1 at 1B and 7B); the faithful-executor conclusion should be replicated on other reasoning-augmented generators (T2I-R1~\cite{t2ir1}, BAGEL~\cite{bagel}) before generalizing across architectures. Validation subsets (300 prompts); headline conditions carry three generation seeds and paired tests, but full-benchmark evaluation remains future work. Detector-based scoring inherits OWLv2's open-vocabulary detection quality; VQA-based color scoring, while validated directionally by the attribute-swap experiment, retains judge biases. The planner-mimic condition substitutes object names into donor chains, whose prose may retain donor-specific details (its slightly lower object-detection rate suggests mild stimulus noise). The color-plan verifier tests color presence, not binding. The attention probe is exploratory and null; mechanism claims await finer-grained analysis. The companion cross-family and encoder-probe measurements referenced in \S\ref{sec:support} predate the audited pipeline and are used only as context, never in direct numerical comparison.

\subsubsection*{Ethics Statement}
Improved compositional control is dual-use: it makes T2I models more useful and makes targeted synthetic imagery easier to produce. Explicit textual plans offer a mitigation surface---harmful objects and relations can be screened at the plan stage before any image exists. Stronger binding fidelity can also render learned social biases more faithfully; planner outputs should be audited for biased associations, not only geometric correctness.

\subsubsection*{Reproducibility Statement}
All experiments use public models and prompt lists, fixed seeds, and stated decode settings (\S\ref{sec:setup}); intervention procedures are fully specified in \S\ref{sec:interventions}--\ref{sec:perturb}. We release the verification, repair, dial, and perturbation scripts, the detector-based plan-fidelity evaluator, all generated plans and images ($\sim$12k), per-image scores, and the paired-statistics analysis code.

\bibliography{references}
\bibliographystyle{splncs04}

\appendix

\section{Per-Relation Planning Accuracy}
\label{app:breakdowns}

\begin{table}[h]
\centering
\small
\begin{tabular}{@{}lcc@{}}
\toprule
Relation & Plan accuracy (original phrasing) & Plan accuracy (inverted phrasing) \\
\midrule
on the left of & 97.8\% & 98.0\% \\
on the right of & 55.1\% & 53.9\% \\
on the top of & 89.5\% & 87.5\% \\
on the bottom of & 73.7\% & 77.9\% \\
\bottomrule
\end{tabular}
\caption{Planner accuracy by relation and mention order (chains only, 3 seeds per prompt). Accuracy follows the relation word regardless of which object is mentioned first---semantically identical layouts differ by up to 44 points depending on phrasing---consistent with a raster-order prior (first-mentioned entity placed at the top-left origin) rather than object-specific or mention-order effects alone.}
\label{tab:perrelation}
\end{table}

\end{document}